\documentclass[arxiv]{melba}

\usepackage{amsmath,amsfonts}

\melbaid{YYYY:NNN}  
\doi{10.59275/j.melba.2026-AAAA}
\melbaauthors{Niyoj Oli and Sachin Acharya}  
\email{niyoj.oli@naamii.org.np}
\volume{2026}
\firstpageno{1}  
\melbayear{2026}  
\datesubmitted{2026-07-24}  
\datepublished{2026-06-19}  

\melbaspecialissue{MICCAI Open Data 2026 x MELBA}
\melbaspecialissueeditors{AT, MS et al.}

\ShortHeadings{SAGE - South Asian GI Endoscopy Dataset}{Oli}

\title{SAGE: An Expert-Annotated South Asian GI Endoscopy Dataset \\for Multimodal Learning and Hallucination Analysis
}

\author{
	\firstname Niyoj \surname Oli\aff{1}\orcid{0009-0009-3660-9030},
	\name Sachin Acharya\aff{1}\orcid{0009-0001-4437-0332}, 
    \name Sandesh Pokhrel$^\dagger$\aff{1,5}\orcid{0009-0001-4843-7899}, 
    \name Sanjay Bhandari$^\dagger$\aff{1,5}\orcid{0009-0009-0722-0739}, 
    \name Ramesh Rana\aff{2}, 
    \name Nikesh Mani Shrestha\aff{2}, 
    \name Ram Bahadur Gurung\aff{2}, 
    \name Yash Raj Shrestha\aff{3}\orcid{0000-0002-2699-4723}, 
    \name Prashnna K Gyawali\aff{4}\orcid{0000-0003-1201-6993},
    \name Binod Bhattarai\aff{1,6}\orcid{0000-0001-7171-6469}
}

\affiliations{
	\num 1 \addr Nepal Applied Mathematics and Informatics  Institute for Research, Nepal\\
	\num 2 \addr Gastrointestinal Department, Dhulikhel Hospital, Nepal\\
    \num 3 \addr University of Lausanne, Switzerland \\
    \num 4 \addr University of West Virginia, USA \\
    \num 5 \addr University of Utah, USA \\
    \num 6 \addr University of Aberdeen, UK 
}

\abstract{
Gastrointestinal cancers represent a growing health burden in the South Asian region, driven largely by rapid changes in socio-economic conditions and lifestyle habits. However, early diagnosis remains limited by inadequate equipment, financial resources, and scarce GI expertise. AI-assisted diagnosis and report generation show great promise in alleviating this problem by providing non-specialist healthcare workers the technical expertise to perform diagnosis. 
Yet, almost all open-source, publicly available datasets are predominantly collected from the European region, with minimal representation from the South Asian region. The lack of open-source GI datasets from diverse geographic regions has made it difficult to assess whether population bias is present in existing models, and to develop geographically inclusive AI tools for automated GI diagnosis.
To address this gap, we introduce SAGE: An Expert-Annotated \underline{S}outh \underline{A}sian \underline{G}I \underline{E}ndoscopy dataset for Multimodal Learning and Hallucination Analysis, for image captioning, multi-label classification, and visual question answering (VQA) tasks. It consists of 1,300 images, their captions along with hallucination tag, 18 labels and 14,276 question-answer pairs making it well-suited for diverse range of tasks including classification, benchmarking, and fine-tuning large multimodal models (LMMs). 
We further conducted benchmarking of task-specific models, like multi-class classifiers on the effect of population shift which reveals that such models suffers the most with an average F1 score drop of 54 points on South Asian dataset. Further, benchmarking of contemporary LMMs reveals a substantial drop in the average GREEN score for anatomical landmark detection (0.308) and abnormality detection (0.410). 
We open-source our dataset under the CC BY-SA 4.0 license, and we hope to encourage others to contribute toward more inclusive dataset representation and help counteract population bias in medical AI. The code is publicly available at \url{https://github.com/bhattarailab/SAGE}, and the dataset at \url{https://www.synapse.org/SAGE}.
}


\begin{document}

\twocolumn[\maketitle]

\section{Background}

\footnotetext{$\dagger$ Work performed while at NAAMII.}

\begin{table*}[t!]
\centering
\caption{Comparison of publicly available gastrointestinal endoscopy 
datasets. SAGE is the only dataset collected from South Asia, 
addressing the critical lack of regional representation in GI imaging 
landscape.}
\label{tab:datasets}
\resizebox{\textwidth}{!}{%
\begin{tabular}{lclc}
    \textbf{Dataset} & \textbf{\#Images} & \textbf{Primary Tasks} & \textbf{Collection Sites} \\
    \hline
    CVC-ClinicDB~\citep{cvcclincdb} & 612 & Polyp segmentation & Spain \\
    Kvasir~\citep{kvasir} & 8,000 & Multi-class classification & Norway\\
    HyperKvasir~\citep{Borgli2020} & 10,662 & Multi-class classification, Segmentation & Norway \\
    Kvasir-Instrument~\citep{kvasirinstrument} & 590 & Instrument segmentation & Norway \\
    GastroVision~\citep{jhagastrovision} & 8,000 & Multi-class classification & Norway, Sweden \\
    REAL-Colon~\citep{Biffi2024} & 2.7M & Region of Interest & Japan, Austria, Italy \\
    Kvasir-VQA~\citep{gautam2024kvasirvqa} & 6,500 & Visual Question Answering & Norway\\
    PolypDB~\citep{polypdb} & 3,934   & Polyp detection (ROI) & Norway, Sweden, Vietnam \\
    GutVLM~\citep{KhaBid_HallucinationAware_MICCAI2025} & 1,816 & Image captioning, VQA & Norway \\
    GastroNet~\citep{gastronet} & 5M & Unlabelled & Netherlands\\
    GastroEndoNet~\citep{BITTO2025111572} & 4,006 & Multi-class classification & Bangladesh \\ 
    GI Bleeding WCE~\citep{khan2023gastrointestinal} & 226 & Multi-class classification & Pakistan \\
    \textbf{SAGE (Ours)} & \textbf{1,300} & \textbf{Image captioning, Multi-label classification, VQA} & \textbf{Nepal}
\end{tabular}}
\end{table*}

	\enluminure{G}{astrointestinal} (GI) cancers account for a substantial proportion of cancer-related deaths in South-East Asia, accounting for 16.9\% of cancer-related deaths in the region. Colorectal cancer ( 9.9\% new cases) and stomach cancer (7.1\%) rank as the second and fourth most incident malignancies in the region \citep{2024cancertoday}. Upper gastrointestinal endoscopy and colonoscopy are minimally invasive procedures for the detection of polyps, adenomas, early malignancy, and other gastrointestinal pathologies. Timely access to these procedures has substantial clinical value, with evidence showing that endoscopic screening is associated with a 40\% relative reduction in gastric cancer mortality~\citep{zhang2018endoscopic}, while colonoscopy screening reduces colorectal cancer incidence by approximately one-fifth~\citep{schoen2012colorectal}. However, in the South Asian region, delayed diagnosis remains prevalent, owing to a persistent shortage of trained gastroenterologists, limited access to modern endoscopic equipment, inadequate diagnostic infrastructure, lack of financial support, and lack of awareness. These factors have collectively contributed to the increasing burden of GI cancers in the South Asian population~\citep{Chandrasinghe2017,pardamean2023changing}.
    
    Recent advances in artificial intelligence (AI) have demonstrated remarkable potential in GI disease assessment and diagnosis, offering a pathway to alleviate this burden through earlier and more accurate detection - even in the presence of imaging artifacts - while reducing dependence on scarce specialists~\citep{anirvan2020ai_gi_endoscopy_resource_constrained}. Despite this promise, the deployment of AI in high-stakes clinical settings, where algorithmic decisions directly impact patient outcomes, remains a subject of considerable debate. A critical and underexplored concern is the bias inherent in existing GI datasets: as shown in Table~\ref{tab:datasets}, contemporary benchmark datasets are predominantly sourced from European institutions, with minimal representation from South Asia. Across geographical boundaries, factors such as endoscopic device quality, bowel preparation standards, disease prevalence and distribution, and image colour characteristics vary considerably. This lack of representational diversity raises fundamental questions about the transferability and clinical applicability of AI-based GI diagnostic tools beyond the populations on which they were trained.
    
    \subsection{Related Works}
    Early efforts to curate GI images began with CVC-ClinicDB \citep{cvcclincdb}, which contains 612 images of single polyps with their corresponding pixel-level segmentation masks. Kvasir~\citep{kvasir} subsequently extended the GI dataset landscape by introducing 8,000 images collected from Norway, oriented toward multi-class classification across 8 categories covering 3 anatomical landmarks and 3 pathological findings. However, its applicability is constrained by the limited number of anatomical and pathological categories. Hyperkvasir~\citep{Borgli2020} further extended this landscape with 10,662 labeled images across 23 class labels, broadening coverage to 6 anatomical landmarks and 12 pathological findings. Despite the increased class diversity, the dataset retains single-site collection bias, raising concerns about generalizability across diverse real world clinical environments.

    GastroVision~\citep{jhagastrovision} tried to address this limitation by introducing the first multi-site GI dataset, collecting 8000 images from two hospitals in Norway and Sweden, covering 27 different classes. REAL-Colon~\citep{Biffi2024} further broadened the geographic coverage by collecting 2.7M colonoscopy frames from sites in Japan, Austria and Italy to target real world generalizability. More recent efforts have focused on specialized, single-task dataset: PolypDB~\citep{polypdb} provides 3,934 images with region-of-interest annotation for polyp detection and Kvasir-Instrument~\citep{kvasirinstrument} offers 590 images for surgical instrument segmentation. However, the datasets are oriented toward classification and segmentation tasks, which are insufficient for training and benchmarking the medical reasoning and language understanding capabilities of modern LMMs.
    
    Kvasir-VQA~\citep{gautam2024kvasirvqa} extended HyperKvasir and Kvasir-Instrument by introducing image captioning and visual question answering (VQA) tasks, enabling AI model development in natural language processing landscape for GI images. The dataset contains 58,800 question-answer pairs spanning 19 question types, including yes/no, color recognition, counting, multiple-choice, and location queries. However, its closed-ended format with a restricted answer space neither adequately exercises the linguistic capabilities nor medical reasoning of modern LMMs. Kvasir-VQA-x1~\citep{Gautam2025Oct} partially addressed this by introducing complex questions requiring multi-step reasoning, enabling evaluation in open-ended settings. Nevertheless, both datasets have limitations in clinical validity: Kvasir-VQA relied on computer scientists with minimal medical knowledge for annotation, while Kvasir-VQA-x1 used LMMs as adjunct annotators rather than clinical experts, raising concerns about the clinical accuracy of the resulting labels. GutVLM~\citep{KhaBid_HallucinationAware_MICCAI2025} extended Kvasir dataset by formulating 12 clinically grounded questions with GI expert involvement and used AI assistance for caption generation, subsequently verified by medical experts. 
    
    Despite these efforts, GI imaging datasets remain heavily concentrated in the European region, raising concerns about population bias and limiting applicability in underrepresented regions such as South Asia. Where South Asian datasets do exist, their scope remains narrow: GastroEndoNet~\citep{BITTO2025111572} was collected from Bangladesh but restricted to GERD and polyp classification, while the GI Bleeding WCE dataset~\citep{khan2023gastrointestinal}, collected from Pakistan, covers only bleeding and lesion classification. Such narrow task coverage limits their suitability for broader range of tasks including benchmarking and finetuning modern LMMs. To address this gap, we introduce SAGE, the first expert annotated GI image captioning and VQA dataset from South Asia dedicated to enabling the development of geographically inclusive LMMs and the systematic assessment of biases present in existing models trained predominantly on GI datasets from European populations.
    
\section{Summary}

    \begin{figure*}[t]
        \centering \includegraphics[width=\linewidth]{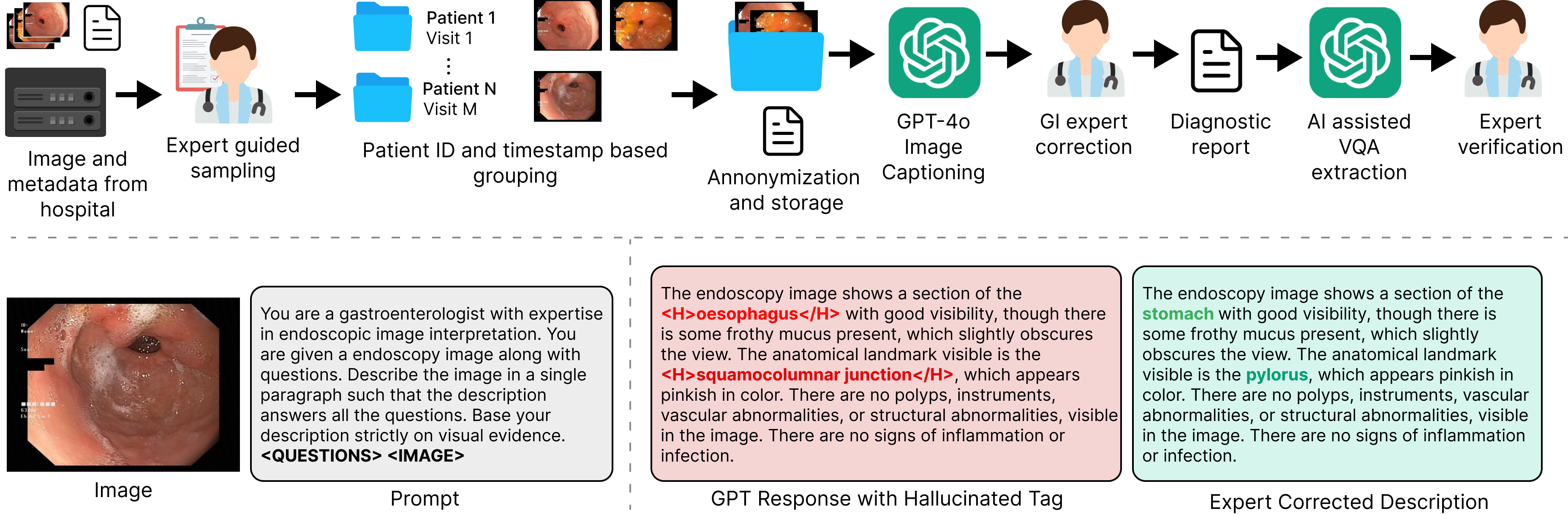}
        \caption{Overview of SAGE dataset annotation pipeline. Top: endoscopy images and associated metadata are collected from hospital, sampled by GI experts, anonymized, and processed through an AI-assisted annotation workflow for description generation and VQA extraction. Bottom left: prompt template used to generate image description from a endoscopy image and predefined questions. The predefined questions are shown in Table~\ref{tab:questions}. Bottom right: example of AI-generated description containing hallucinated findings highlighted in red and the corresponding GI expert correction shown in green. \texttt{<H>}...\texttt{</H>} denotes hallucinated markers.}
        \label{fig:methodology}
    \end{figure*}

    SAGE addresses a critical representation gap in the GI endoscopy dataset landscape by introducing the first expert-annotated South Asian GI imaging dataset to support such a broad range of tasks including image captioning, VQA, multi-label classification, and hallucination-aware fine-tuning. Further, the study also benchmarks the performance of five contemporary LMMs including Qwen 2.5 VL 72b, Gemini-3 Flash Preview,  Gemma 4 31B, Grok 4.3, and Claude Sonnet 4.6 on GI image captioning tasks and additionally conducts experiments to study geographical bias in model performance.

\section{Discussion}

\begin{table}[h]
    \caption{Clinical categories and summarized prompt questions used to guide GPT-based AI-assisted annotation of endoscopy images. For brevity and due to manuscript space constraints, only condensed versions of the questions are shown. The complete prompt template is available in the code repository.}
    \centering
    \begin{tabular}{p{0.42\columnwidth} p{0.45\columnwidth}}
        \textbf{Category} & \textbf{Question Summary} \\
        \hline
        Visibility & Quality of view; presence and type of obstruction \\
        Section identification & Whether a specific section of GI tract is identifiable \\
        Anatomical landmarks & Presence, description, and color of landmarks \\
        Polyps & Count, location, and PARIS classification \\
        Instruments & Name, location, action, and target of instruments \\
        Vascular abnormalities & Name, color, and position \\
        Structural abnormalities & Name, color, and position \\
        Growth abnormalities & Name, color, and position \\
        Other abnormalities & Name, color, and position \\
        Inflammation & Name, color, and position \\
        Infection & Name, color, and position \\
        Special findings & Description \\
    \end{tabular}
    \label{tab:questions}
\end{table}
    SAGE consists of 1,300 de-identified GI images that are collected from Dhulikhel Hospital, Nepal along with description answering 12 GI expert curated questions, 14,276 question-answer pairs, multi-label categories spanning 18 distinct classes, as illustrated in Table \ref{tab:data-details}. The dataset further includes GPT-generated descriptions for each GI image, accompanied by expert-curated tags identifying hallucinated content and their corresponding corrections. The dataset includes annotation in JSON format, metadata in CSV format, and images in JPG to facilitate discovery and downstream use. This dataset has been prepared in accordance with the FAIR principles to support findability, accessibility, interoperability, and reusability~\citep{wilkinson2016fair}.

    Due to resource constraints, only 1,300 images could be collected and annotated for this initial release. Additionally, the collection site lacked video recording and storage capability for full endoscopic procedures; consequently, the available images are limited to those that the performing gastroenterologist deemed necessary to save. We attempted to mitigate this selection bias through expert review; however, residual selection bias introduced by the performing gastroenterologist may still persist in the dataset. This infrastructure limitation also resulted in a restricted number of high-visibility images for certain classes, such as mucosal growths and bulges, and certain anatomical landmarks, most notably the landmarks of lower GI.

    Furthermore, as the study was retrospective in nature and the collection site lacked an effective metadata storage system, demographic details (age, gender), as well as procedure metadata such as the device used, are unavailable for a significant proportion of records. Reported summary statistics for age and gender distribution are therefore derived from the subset of patients with available metadata, representing approximately 56.06\% of the patients.
    
    To address these limitations, we plan a future multi-site prospective study to collect GI images while prioritizing systematic collection of procedure metadata, patient demographics, and coverage of a broader range of anatomical landmarks and pathological findings.

    Independent of these future directions, users are encouraged to treat model outputs as supplementary decision-support rather than a substitute for expert clinical judgment, and to ensure proper attribution and sharing of any derivative works under the same license. Furthermore, users must ensure the ethical handling of the data, refraining from any attempt at re-identification or misuse of sensitive information.

\section{Resource Availability}
    \subsection{Data/Code Location}
	The dataset is publicly available on Synapse (DOI: 10.730\\3/syn75397327)~\citep{Oli_Acharya_Pokhrel_Bhandari_Rana_Shrestha_Gurung_Shrestha_Gyawali_Bhattarai_2026} at \url{https://www.synapse.org/SAGE}. The code used for dataset development and validation experiments is available at \url{https://github.com/bhattarailab/SAGE}.

    \subsection{Potential Use Cases}
	SAGE provides expert-annotated dataset for fine-tuning LMMs on GI image captioning and VQA, extending LMM capabilities to the GI imaging domain. Further, the hallucination span tags for GPT-generated responses, together with the expert-annotated corrections in our dataset, enable hallucination-aware fine-tuning~\citep{KhaBid_HallucinationAware_MICCAI2025} of LMMs. This approach has demonstrated notable improvements in response quality, as models learn to generate accurate clinical descriptions while also developing explicit awareness of hallucination patterns. The dataset additionally provides class labels for multi-label classification, usable to train and evaluate deep learning models, supporting robust GI image classifiers and generalizable vision encoders.

    Clinically, the breadth of annotations and geographic diversity in SAGE can support development of more accurate, robust LMMs for real-world deployment in underrepresented settings such as South Asia, where AI-assisted GI screening may help address resource constraints and improve access to timely diagnostic support. Furthermore, as the first expert-annotated South Asian GI endoscopy dataset supporting image captioning, VQA, and multi-label classification, SAGE provides a structured benchmark for evaluating geographic bias in contemporary AI models, addressing a longstanding gap in assessing model performance beyond European populations.

    \subsection{Licensing}
	This dataset is released under the Creative Commons Attribution ShareAlike 4.0 International License (CC BY-SA 4.0)\footnote{https://creativecommons.org/licenses/by-sa/4.0/}. Users may share and adapt the dataset, including for commercial purposes, provided that appropriate credit is given and any adapted material is distributed under the same or a compatible license.

    

\section{Methods}
    \subsection{Data Details}

\begin{figure*}[t]
    \centering
    \includegraphics[width=\linewidth]{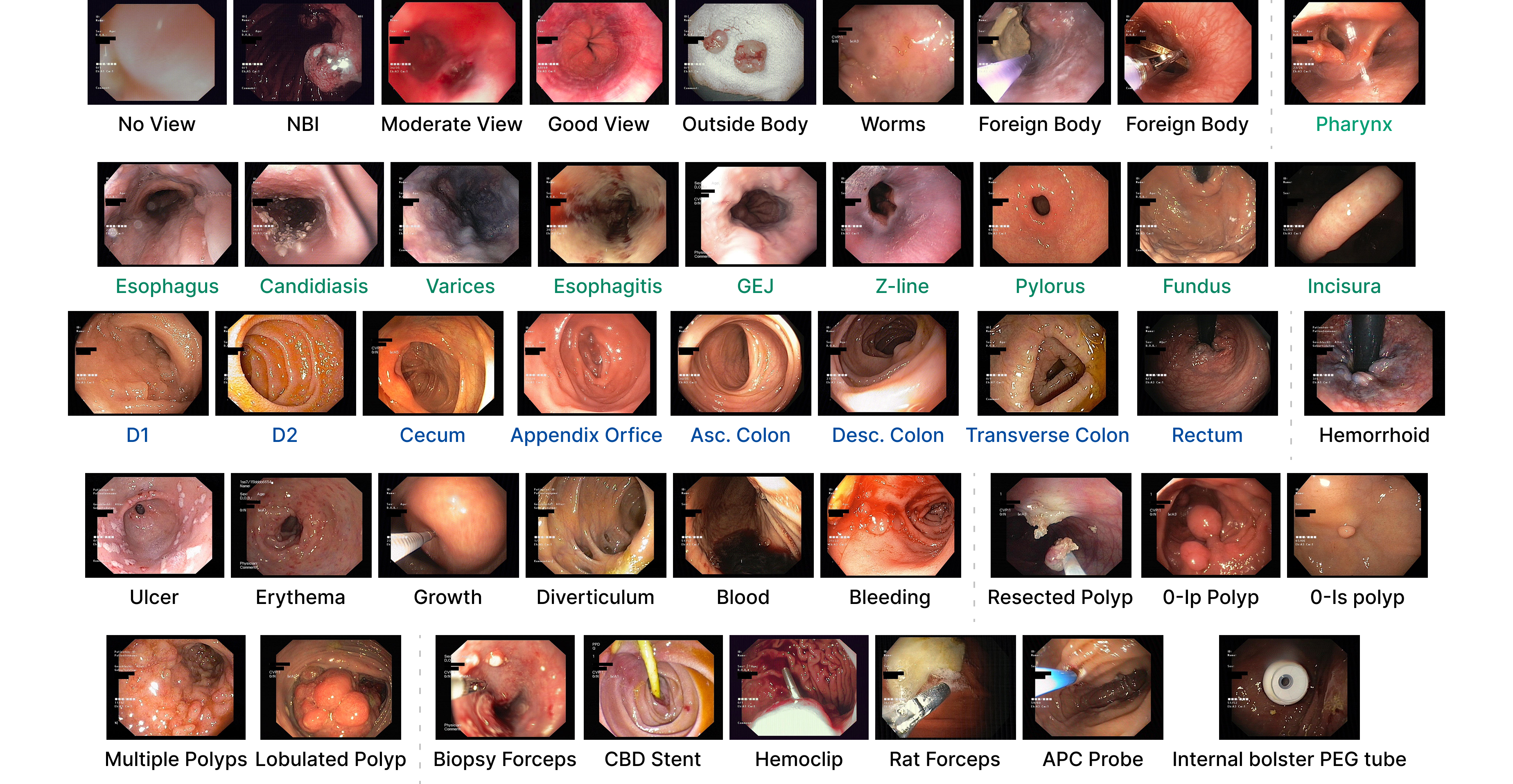}
    \caption{Example images from SAGE dataset illustrating anatomical landmarks, gastrointestinal segments, instruments, and abnormalities. Images with green labels correspond to upper gastrointestinal (GI) tract images, whereas those with blue labels correspond to lower GI tract images. Asc. colon and Desc. colon denote the ascending and descending colon, respectively.}
    \label{fig:samples}
\end{figure*}
\begin{figure*}[h]
    \centering
    \includegraphics[width=1\linewidth]{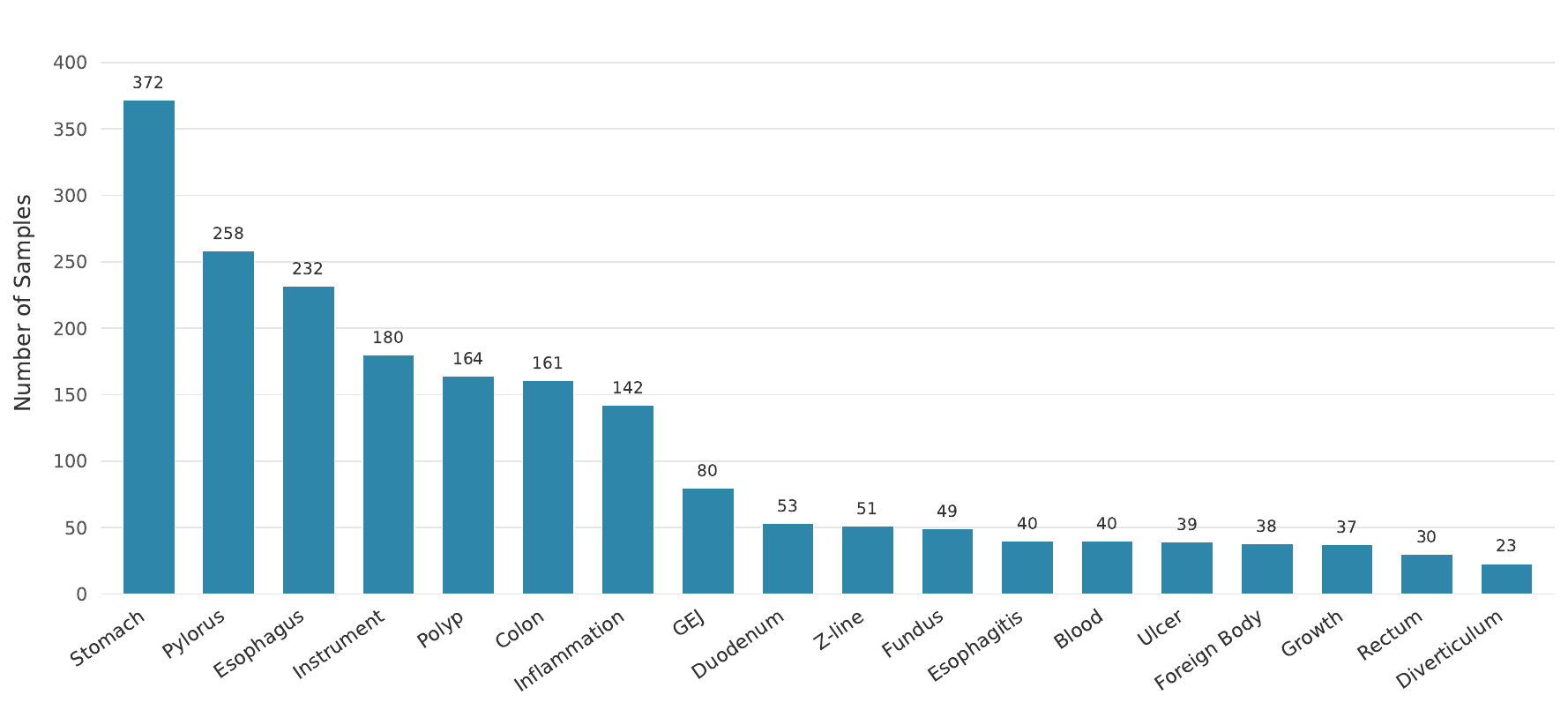}
    \caption{Distribution of annotated frames across 18 classes in the multi-label gastrointestinal endoscopy dataset.}
    \label{fig:distribution}
\end{figure*}

    Table~\ref{tab:data-details} presents detailed dataset specifications, Figure~\ref{fig:distribution} shows per-label sample counts, and Figure~\ref{fig:samples} illustrates the image types (anatomical landmarks, foreign bodies, luminal findings, polyps, and instruments). 
    
\begin{table}[h]
    \caption{SAGE dataset composition and specification. $^\dagger$Demographic statistics are derived from the subset of patients with available metadata (56.06\% of the patients).}
    \label{tab:data-details}
    \centering
    \begin{tabular}{lc}
         \textbf{Characteristics} &  \textbf{Value} \\
         \hline
         \textbf{Image Properties} & \\
         Total images & 1,300 \\
         Image resolution & 768px $\times$ 576px \\
         Image format & JPG \\
         \hline
         \textbf{Collection} & \\
         Patients & 290 \\
         Visits & 371 \\
         Average images per patient & 4.48 $\pm$ 3.55 \\
         Average images per visit & 3.50 $\pm$ 1.96 \\
         \hline
         \textbf{Age$^\dagger$} & \\
         Mean $\pm$ SD & 52.35 $\pm$ 17.36 \\
         Median [Min, Max] & 51 [3, 88]\\
         \hline
         \textbf{Gender$^\dagger$ (in \%)} & \\
         Male & 46.63\%\\
         Female & 52.76\%\\
         \hline
         \textbf{Annotation} & \\
         QA Pairs & 14,276\\
         Unique Classes & \\
         \quad Anatomical landmarks & 5\\
         \quad Section & 4\\
         \quad Luminal findings & 8\\
         \quad Instrument & 1 \\
    \end{tabular}
\end{table}

    \subsection{Methods Used for the Data Creation}
	SAGE was constructed through a multi-stage pipeline ensuring clinical relevance, patient privacy, and accurate multimodal annotations. Figure \ref{fig:methodology} overviews these steps, which are detailed in the following subsections.
    
    \subsubsection{Data Acquisition}
    This study was retrospective in design; thus, images were collected from the hospital's endoscopy database. Prior to data acquisition, a list of predefined target findings, based on their prevalence in South Asia, was established by the gastroenterology (GI) experts within the research team. This list served as the basis for manually screening patients from the procedure logbook entries. 
    To ensure that the dataset comprised clinically significant images 
    two senior gastroenterologists curated a representative image selection for each patient. Images containing duplicates, near-duplicates, motion blur, or out-of-focus frames were excluded, with preference given to images demonstrating pathological findings or distinct anatomical landmarks.

    In addition to images, metadata with patient age (in years), gender, and the endoscopy device used were collected. However, due to infrastructure constraints, complete metadata collection was not possible for all cases; device information was available for only 226 of the images (17.38\%), while age and gender information were missing for a subset of cases. Among the recorded devices, the Olympus GIF-1TQ160 was the most frequently used model (67 images), followed by the Olympus GIF-Q165 (39), Olympus CF-Q165L (38), Olympus CF-Q145L (31), Olympus CF-Q165I (28), Olympus GF-160 (9), CF-H180A1 (6), GIF-HQ190 (5), and Olympus GIF-H185 (3). Descriptive statistics for all available metadata are presented in Table~\ref{tab:data-details}.
    
    \subsubsection{Anonymization Protocol}
   The GI images in the hospital's database were originally stored in JPG format. To anonymize them, only the RGB pixel data were extracted, producing new image objects free of the original EXIF and other ancillary metadata. However, because the image capture timestamp had been overlaid directly onto the pixel data rather than stored as metadata, it remained visible in the copied images and required separate handling.
   
    For patient-level anonymization, images were grouped by patient and visit, with each image assigned a randomly generated patient UUID and visit UUID; no mapping between these and the original patient records was retained. Visit dates were converted to relative timestamps, with the first image in each visit designated as t = 0, and all visible dates subsequently redacted from the images. All images were then manually verified to confirm completeness and accuracy of the redaction. Through this process, all images were anonymized while retaining key clinical metadata, including patient gender, age at the time of procedure, device information, and relative timestamp for each sample.
    
    \subsubsection{AI Assisted Annotation}
    The anonymized images were each passed to GPT-4o~\citep{openai2024gpt4technicalreport} along with a structured prompt comprising 12 questions prepared by GI experts, as illustrated in Figure~\ref{fig:methodology}. As the GPT-generated descriptions may contain hallucinated information, each output was reviewed and corrected by a GI expert. Hallucinated spans and their corresponding corrections were tagged and stored, and has been released alongside the dataset. The corrected diagnostic description was subsequently passed to GPT-4o-mini to structure the content into 12 question-answer pairs, with non-applicable questions removed during this step. GPT-4o-mini was selected for this step owing to its lower cost and the comparatively reduced complexity of the question-answer extraction task relative to the initial description generation. The resulting question-answer pairs were further evaluated by a GI expert to ensure clinical accuracy. All annotations were performed by two medical experts using our in-house annotation software, and subsequently reviewed by three senior gastroenterologists from Dhulikhel Hospital.

    Similarly, GI experts on our team reviewed the corrected description of each image to assign the applicable label(s) from a set of 18 pre-defined classes. Patient-level classes were then obtained by taking the union of classes across all images belonging to a given patient. This patient-level class set was used to perform multi-class stratified sampling, splitting patients into train (70\%) and test (30\%) sets. This approach prevented patient-level train-test leakage while preserving class distribution across the two splits.

\section{Validation}
    \subsection{Inter-Rater Disagreement}
    To validate the agreement between annotators, inter-rater agreement was measured using Krippendorff's $\alpha$ which accommodates multiple raters, missing data, and supports different levels of measurement ~\citep{krippendorff2024}. A random sample of 100 images was drawn from the dataset and independently annotated by three medical expert annotators. Inter-rater disagreement was computed across six clinically relevant categories: visibility, section identification, landmark identification, polyp detection and counting, instrument detection, and abnormality detection. Krippendorff's $\alpha$ was computed using the Python \texttt{krippendorff} library~\citep{castro-2017-fast-krippendorff}. The coefficient ranges from $-1$ to $+1$, where $-1$ indicates systematic disagreement, $0$ indicates agreement at chance level, and $+1$ represents perfect agreement. Results are reported in Table~\ref{tab:interrater}.

\begin{table}[h]
    \centering
    \caption{Inter-rater agreement in endoscopic image captioning across 
    clinically relevant categories, measured using Krippendorff's $\alpha$.}
    \begin{tabular}{lccl}
        \textbf{Category} & \textbf{Scale} & \textbf{$\alpha$} & \textbf{Examples} \\
        \hline
        Visibility & Ordinal & 0.5803 & \{good, fair, no-view\} \\
        Section & Nominal & 0.5309 & \{stomach, colon, \dots\} \\
        Landmark & Nominal & 0.5123 & \{GEJ, pylorus, \dots\}\\
        Instrument & Nominal & 0.6945 & \{snare, forceps, \dots\}\\
        Abnormality & Nominal & 0.6211 & \{ulcer, blood, \dots\}\\
        Polyp Count & Ratio & 0.7571 & [0, $\infty$] \\
    \end{tabular}
    \label{tab:interrater}
\end{table}

    Instrument detection ($\alpha = 0.694$) and polyp count ($\alpha = 0.757$) both exceed the threshold of $0.667$, which Krippendorff defines as the lower bound for drawing tentative conclusions~\citep{krippendorff2024}. Abnormality detection ($\alpha = 0.621$) and visibility ($\alpha = 0.580$) approach this threshold, reflecting the inherent perceptual variability in assessing endoscopic findings. The comparatively lower agreement for section identification ($\alpha = 0.531$) and landmark identification ($\alpha = 0.512$) is expected given the morphological complexity and anatomical ambiguity involved in gastrointestinal tract assessment~\citep{vanDoorn}. These values are consistent with inter-rater agreement levels reported in the endoscopy literature~\citep{vanDoorn, intrapapilary}.
    
    \subsection{Baseline Performance}
    We performed multi-label classification experiments on SAGE dataset as a baseline for future experiments. We trained ResNet-50 and DenseNet-121 on SAGE dataset and used mean average precision (mAP) with weighted, macro, and micro averaging as the evaluation metrics. Weighted and micro averaging considers the support for each class while macro averaging treats all the class equally.  
    
    The baseline results are reported in Table~\ref{tab:baseline}. The DenseNet model consistently outperforms ResNet in all the evaluation metrics. However, both of the model suffer from class imbalance evident by the difference in weighted average and macro average scores (weighted mAP of DenseNet = 0.4226 and macro mAP = 0.3041). We further used ResNet and DenseNet, pretrained on the HyperKvasir dataset, represented by ResNet-50$^H$ and DenseNet-121$^H$, and fine-tuned them on SAGE dataset. This pretraining boosted the performance of both models across all the metrics, where ResNet$^H$ macro mAP increases from 0.1819 to 0.2154 whereas for DenseNet$^H$ macro mAP increases by 0.0921.

\begin{table}[h]
    \centering
    \caption{Baseline multi-label classification performance (mAP). $^H$ denotes models 
    initialized with weights pretrained on HyperKvasir, with the classification head 
    re-initialized prior to fine-tuning on the target dataset.}
    \begin{tabular}{lccc}
    \textbf{Model} & \textbf{Weighted} & \textbf{Macro} & \textbf{Micro} \\
    \hline
    ResNet-50 & 0.3130 & 0.1819 & 0.3098 \\
    DenseNet-121 & 0.4426 & 0.3041 & 0.4580 \\
    ResNet-50$^H$ & 0.5128 & \textbf{0.3973} & 0.5177\\
    DenseNet-121$^H$ & \textbf{0.5146} & 0.3962 & \textbf{0.5334} \\
    \end{tabular}
    \label{tab:baseline}
\end{table}

    \subsection{Effects of Population Shift}
    The consequences of population shift in AI models, especially for GI imaging, remain an underexplored area of study. To investigate this effect, we trained multi-class classification models separately on two European datasets - HyperKvasir and GastroVision - and evaluated them on SAGE dataset from South Asia. Because the label spaces differed between the datasets, those of HyperKvasir and GastroVision were mapped onto SAGE's label space, considering class hierarchy and GI expert judgment. Samples with classes not present in SAGE were removed. The complete experimental details are available in our code repository.

    The results are reported in Table~\ref{tab:population-shft}, which shows that the performance of models trained on European datasets decreases substantially when evaluated on the South Asian dataset. For HyperKvasir-trained models, the F1 score drops by 0.7543 for ResNet and by 0.6691 for DenseNet; a similar trend is observed for the GastroVision-trained models. This highlights the need for more geographically diverse and inclusive datasets, which would aid the deployment of health AI models in low and middle income countries.

\begin{table}[h]
    \centering
    \caption{In-domain vs. external (ours) performance of ResNet-50 and DenseNet-121 trained on HyperKvasir and GastroVision, illustrating the impact of population shift. F1$^{ID}$ and F$^{SAGE}$ represent the F1 score on their own test set and on SAGE dataset, respectively.}
    \begin{tabular}{llcc}
         \textbf{Dataset} & \textbf{Model} & \textbf{F1$^{ID}$} & \textbf{F1$^{SAGE}$}  \\
         \hline
         Hyperkvasir & ResNet-50 &  0.8707 & 0.1164 \\
                     & DenseNet-121 & 0.8382 & 0.1691 \\
         \hline
         Gastrovision & ResNet-50 &  0.5738 & 0.1265 \\
                      & DenseNet-121 &  0.5650 & 0.2431 \\
    \end{tabular}
    \label{tab:population-shft}
\end{table}

    \subsection{Benchmarking Contemporary LMMs}
\begin{figure}[h]
    \centering
    \includegraphics[width=\linewidth]{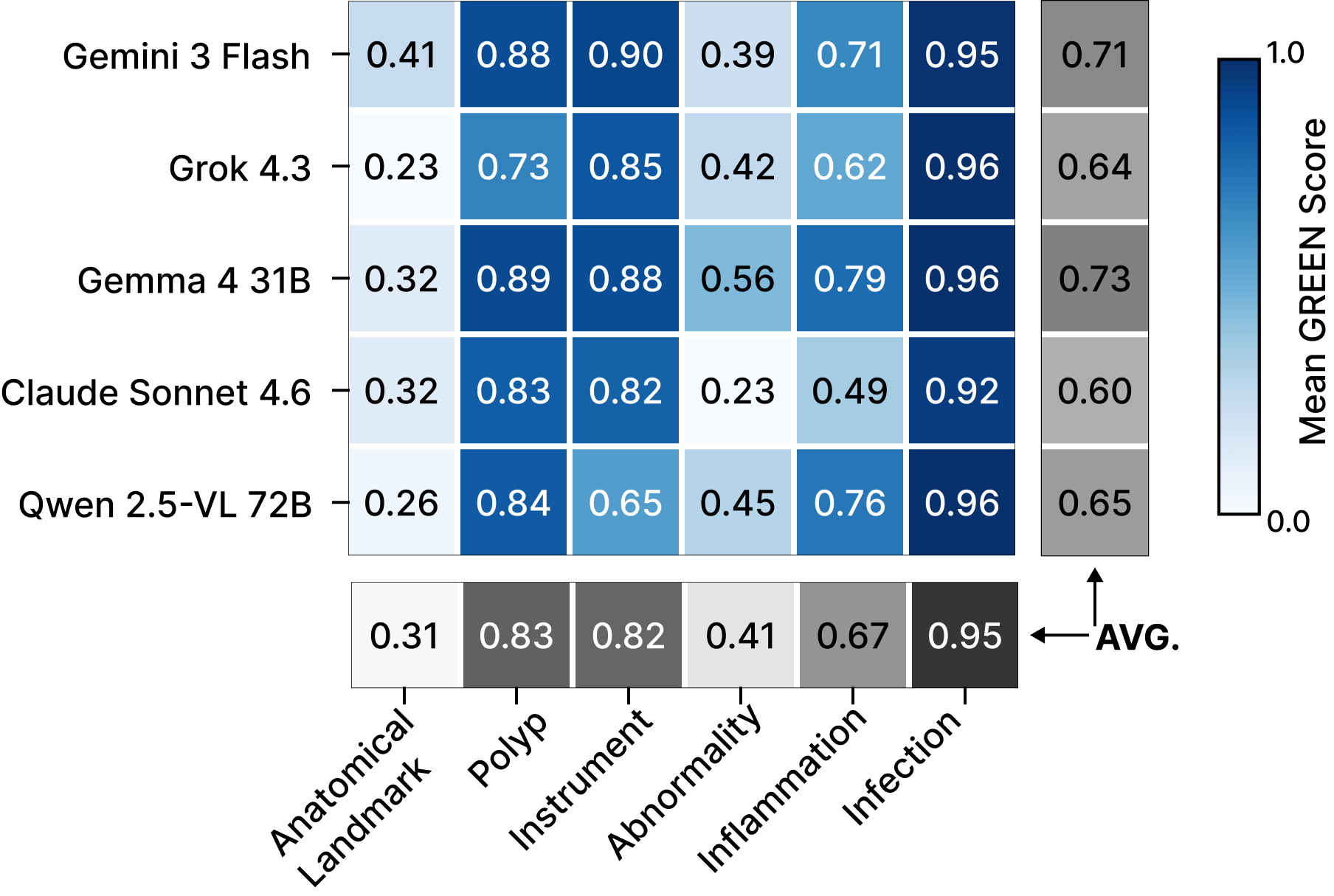}
    \caption{GREEN score of contemporary LMMs across six clinically relevant task. Higher is better. Average values are denoted by grayscale cells.}
    \label{fig:heatmap}
\end{figure}

     Having established that task-specific classifiers trained on European data degrade sharply under population shift, we ask whether general-purpose LMMs inherit the same fragility and how reliably they perform on South Asian GI endoscopy. We benchmark five contemporary LMMs on SAGE dataset: each model generates a caption per image, which is converted into QA pairs for fine-grained evaluation across six clinically relevant tasks and scored against ground truth using the GREEN model~\citep{ostmeier-etal-2024-green}, where a higher score indicates closer clinical agreement. Overall per-model scores and per-category breakdown are reported in Figure~\ref{fig:heatmap}.
     
    Unlike the narrow classifiers, contemporary LMMs remains comparatively robust overall with Gemma 4 31B the strongest at mean GREEN score of 0.73. This aggregate robustness, however, masks where the models fail. The per-category scores in the Figure~\ref{fig:heatmap} show performance collapsing on abnormality detection which falls to low as 0.23 and on landmark identification, which ranges from only 0.23 to 0.41 across all the models, even as some models score up to 0.96 on infection.
    
    Failure on abnormality detection and anatomical landmark identification is especially concerning, as these tasks require GI-specific clinical knowledge and GI scene understanding capability. This weakness is shared by every model rather than confined to one, a consistency that points to a limitation of current LMMs on this dataset rather than a quirk of any single system. High aggregate scores therefore offer little assurance of clinical safety: a model can caption common findings well while still failing on the cases that matter most. Closing this gap calls for geographically representative, expert-annotated data such as SAGE, not as an out-of-distribution test set, but integrated into the training and evaluation of LMMs.

\acks{The work was funded through Open Data MICCAI 2026 grant. The authors acknowledges the support of Gastrointestinal Department team at Dhulikhel Hospital during the data collection process.}

\ethics{This study was approved by the Institutional Review Committee, Kathmandu University School of Medical Sciences, Dhulikhel Hospital (Approval No. 255/25). As the study was retrospective in nature and did not involve the collection of any personal health information (PHI), a waiver of informed consent was granted by the Institutional Review Board. All data were anonymized prior to release, in accordance with applicable ethical standards and regulations regarding human subjects, and no mapping between the randomized patient identifiers in the dataset and patient identities is retained by either the hospital or the research team. For any data-related or ethical inquiries, please contact \url{niyoj.oli@naamii.org.np}.}

\coi{The authors declare that they have no conflicts of interest.}

\data{The dataset is publicly available on Synapse (DOI: 10.730\\3/syn75397327)~\citep{Oli_Acharya_Pokhrel_Bhandari_Rana_Shrestha_Gurung_Shrestha_Gyawali_Bhattarai_2026} at \url{https://www.synapse.org/SAGE}.} 

\bibliography{sample}

\end{document}